%% file: tmi.tex
\documentclass[journal,twoside,web]{ieeecolor}
\usepackage{tmi}
\usepackage{cite}
\usepackage{amsmath,amssymb,amsfonts}
\usepackage{graphicx}
\usepackage{textcomp}
\usepackage{hyperref}

\usepackage{threeparttable}
\usepackage{multirow}
\usepackage{tablefootnote}
\usepackage{xcolor}
\usepackage{algpseudocode}
\usepackage{algorithm}

\newcommand{\ie}{\textit{i}.\textit{e}.}
\newcommand{\eg}{\textit{e}.\textit{g}.}
\newcommand{\bfsection}[1]{\noindent\textbf{#1.}}

\input{math_commands.tex}

\algnewcommand\algorithmicnot{\textbf{not}}
\algdef{SE}[IF]{IfNot}{EndIf}[1]{\algorithmicif\ \algorithmicnot\ #1\ \algorithmicthen}{\algorithmicend\ \algorithmicif}%

\def\BibTeX{{\rm B\kern-.05em{\sc i\kern-.025em b}\kern-.08em
    T\kern-.1667em\lower.7ex\hbox{E}\kern-.125emX}}
\begin{document}
\title{TriSAM: Tri-Plane SAM for zero-shot cortical blood vessel segmentation in VEM images}
\author{Jia Wan, Wanhua Li, Jason Ken Adhinarta, Atmadeep Banerjee, Evelina Sjostedt, Jingpeng Wu, \\ Jeff Lichtman, Hanspeter Pfister, Donglai Wei
\thanks{}
\thanks{Jia Wan is with the School of Computer Science and Technology,
Harbin Institute of Technology (Shenzhen) (e-mail: jiawan1998@gmail.com). }
\thanks{Wanhua Li and Hanspeter Pfister are with the School of Engineering and Applied Sciences,
Harvard University  (e-mail:
\{wanhua,pfister\}@seas.harvard.edu).}
\thanks{Jason Ken Adhinarta, Atmadeep Banerjee, and Donglai Wei are with the Department of Computer Science,
Boston College (e-mail: \{jason.adhinarta,donglai.wei\}@bc.edu).}
\thanks{Evelina Sjostedt and Jeff Lichtman are with the
Department of Molecular and Cellular Biology,
Harvard University (e-mail: \{evelina\_sjostedt@fas.harvard.edu, jlichtman@mcb.harvard.edu).}
\thanks{Jingpeng Wu is with the Lin Gang Laboratory (e-mail: jingpengw@lglab.ac.cn)}
}

\maketitle

\begin{abstract}
While imaging techniques at macro and mesoscales have garnered substantial attention and resources, microscale Volume Electron Microscopy (vEM) imaging, capable of revealing intricate vascular details, has lacked the necessary benchmarking infrastructure. 
In this paper, we address a significant gap in this field of neuroimaging by introducing the first-in-class public benchmark, \textbf{BvEM}, designed specifically for cortical blood vessel segmentation in vEM images. 
Our BvEM benchmark is based on vEM image volumes from three mammals: adult mouse, macaque, and human. We standardized the resolution, addressed imaging variations, and meticulously annotated blood vessels through semi-automatic, manual, and quality control processes, ensuring high-quality 3D segmentation.
Furthermore, we developed a zero-shot cortical blood vessel segmentation method named TriSAM, which leverages the powerful segmentation model SAM for 3D segmentation. To extend SAM from 2D to 3D volume segmentation, TriSAM employs a multi-seed tracking framework, leveraging the reliability of certain image planes for tracking while using others to identify potential turning points. This approach effectively achieves long-term 3D blood vessel segmentation without model training or fine-tuning. Experimental results show that TriSAM achieved superior performances on the BvEM benchmark across three species. Our dataset, code, and model are available online at
\url{https://jia-wan.github.io/bvem}.
\end{abstract}


\begin{figure}[t]%
\centering
\includegraphics[width=\linewidth]{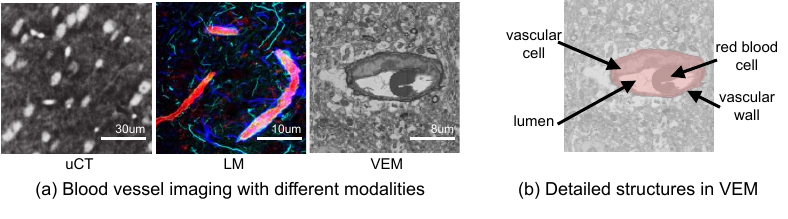}%
\caption{Imaging modalities for blood vessel analysis. (a) Both microtomography ($\mu$CT)~\cite{dyer2017quantifying} and light microscopy (LM)~\cite{lochhead2023high} can only capture blood vessels in the cortex at the sub-micron resolution without ultrastructure details. (b) Volume electron microscopy (VEM) can show unbiased details of the vasculature including all types of cells at a higher resolution. }
\label{fig:vem}%
\end{figure}
\section{Introduction}
\label{intro}
With around 2\% of body weight, our brain receives around 20\% of blood supply. 
Most of the energy and nutrients are consumed by the neurons, and neuron function is sensitive to the blood supply~\cite{andreone2015neuronal, peters2004selfish}. The blood supply can even be adjusted following the consumption of brain regions, called neurovascular coupling~\cite{iadecola2004neurovascular}. 
Alterations of blood vessel structures are observed in many brain diseases, \eg, Alzheimer's and vascular dementia~\cite{kalaria2010vascular}. 
Thus, blood vessels in the brain have been extensively investigated with various imaging modalities at different resolutions (Figure~\ref{fig:vem}a). 
Compared to the macro-level imaging (\eg, CT~\cite{dyer2017quantifying} , MRI~\cite{mcdonald2003imaging}) and mesoscale-level imaging (\eg, light microscopy~\cite{lochhead2023high}), volume electron microscopy (VEM)~\cite{peddie2022volume} can further reveal the detailed ultrastructure including all vascular cells (Figure~\ref{fig:vem}b) for in-depth analysis. 
However, no large-scale annotated VEM dataset exists to develop and evaluate automated 3D blood vessel segmentation methods.

Traditionally, the imaging methods at the macro and mesoscale are widely used and have produced a large amount of data, and a variety of image segmentation algorithms, public datasets, and evaluation methods have been developed ~\cite{goni2022brain,moccia2018blood}. 
At the microscale level, the sample size of VEM is normally limited and most image analyses focus on neuron reconstruction, and blood vessels are largely ignored. 
Recently, owing to the rapid improvement of imaging technology, the sample size of VEM is significantly increased covering all the layers of the cerebral cortex of mouse~\cite{microns-phase2}and human brain~\cite{shapson2021connectomic}, as well as the whole brain of fly~\cite{flywire}. 
Moreover, imaging the whole mouse brain using VEM technology is under planning~\cite{abbott2020mind}.

Thus, we first curate the \textbf{BvEM dataset}, the largest-to-date public benchmark dataset for cortical blood vessel segmentation in VEM images to foster segmentation method development.  
The raw image volumes are from recent publications, which are the largest for each of the three mammals: mouse, macaque, and human acquired at different VEM facilities. 
We downsampled the volumes to a consistent resolution and performed extensive blood vessel annotation, including manual proofreading, semi-automatic segmentation error correction, and quality control, involving multiple rounds of scrutiny by neuroscience experts to ensure accuracy and completeness.
However, on the BvEM dataset, existing 3D blood vessel segmentation methods suffer from two major challenges: the diversity of the image appearance due to variations in the imaging pipeline and the complexity of the 3D blood vessel morphology. Conventional blood vessel segmentation heavily depends on a substantial volume of manually annotated data, a resource that is notably scarce in the existing literature.

To address these challenges, we propose a zero-shot 3D segmentation method, \textbf{TriSAM}, 
leveraging the recent 2D segmentation foundation model, Segment Anything Model (SAM)~\cite{kirillov2023segment}, to handle the appearance diversity. 
To re-purpose SAM for 3D segmentation, we developed a multi-seed segment tracking framework based upon the video object segmentation method with SAM~\cite{yang2023track}.
As it is easier to track 2D blood vessel segments along the blood flow direction, TriSAM selects the best 2D plane for SAM-based tracking and further introduces a recursive seed sampling strategy that performs tri-plane selection at the potential turning points for efficiency.
{The proposed method is similar to Flood-Filling Networks (FFN) \cite{januszewski2018high} by extending the mask in 3D space. Instead of training a 3D neural network, we utilize SAM for prediction since it generalize better. Second, FFN densely extends to 3D while the proposed method extends along the blood vessels. Finally, FFN uses only one seed while we are using growing seeds for long-term tracking.}
Experimental results demonstrate that the proposed TriSAM method significantly outperforms the prior state-of-the-art methods on the proposed BvEM benchmark across all three species.

\section{Related Work}
\label{related}
\bfsection{Blood vessel segmentation methods}
Most existing VEM image segmentation algorithms were developed for neurons~\cite{lee2019convolutional} and synapses~\cite{wu2023out,turner2020synaptic,buhmann2021automatic}. The traditional VEM dataset size is too small to study blood vessel architecture. Recently, with rapid technology improvement, VEM sample size reached the cubic millimeter scale covering all the layers of cerebral cortex~\cite{shapson2021connectomic,microns-phase2}, providing rich details of vasculature structure at scale. However, due to the lack of efficient and effective automatic methods, the blood vessels were segmented manually in the human cortex dataset~\cite{shapson2021connectomic}.
On another front, automatic blood vessel segmentation methods have been developed for other image modalities, such as light microscopy and MRI, at a lower resolution.
We refer to ~\cite{goni2022brain,moccia2018blood} for detailed reviews on filter-based methods~\cite{tsai2009correlations,ji2021brain} and deep learning methods~\cite{tetteh2020deepvesselnet,livne2019u}. Due to the burden of annotation, efforts have been made to decrease the need for annotations~\cite{dang2022vessel}. 
In this paper, we propose the first VEM cortical blood vessel benchmark to foster novel methods.


\bfsection{Segment anything-based methods} As a foundation model for image segmentation, the recently proposed Segment Anything Model~\cite{kirillov2023segment} has garnered significant attention and has been extended to a variety of domains~\cite{cheng2023segment,yu2023inpaint,ma2023segment,deng2023segment} including object tracking~\cite{cheng2023segment,yang2023track}, image inpainting~\cite{yu2023inpaint}, image mattting~\cite{yao2023matte}, super-resolution~\cite{lu2023can}, 3D point cloud~\cite{liu2023segment}, and image editting~\cite{gao2023editanything}.  
Despite SAM's remarkable generalization capabilities, it still encounters some challenges in practical applications,  
One of these challenges is the huge computation costs due to the heavyweight image encoder. FastSAM~\cite{zhao2023fast} adopted a conventional CNN detector with an instance segmentation branch for the segment anything task with real-time speed. MobileSAM~\cite{zhang2023faster} proposed decoupled distillation to obtain a small image encoder, which achieved approximately five times faster speed compared to FastSAM while also being seven times smaller in size. Therefore the MobileSAM is employed in our proposed method. One important challenge lies in the unsatisfactory performance of SAM when confronted with special domains, such as medical~\cite{chen2023sam} or biological~\cite{archit2023segment} images, particularly in the context of 3D data. 
Although SAM demonstrates impressive generalization capabilities, it faces challenges in specialized domains like medical~\cite{chen2023sam,gong20233dsam} or biological images~\cite{archit2023segment}, especially in handling 3D data.
\cite{deng2023segment} assessed the SAM model's zero-shot segmentation performance in the context of digital pathology and showed scenarios where SAM encounters difficulties. 
\cite{mazurowski2023segment} extensively evaluates the SAM for medical image segmentation across 19 diverse datasets, highlighting SAM's performance variability.
To address the domain gap between natural and medical images, SAM-Adapter~\cite{chen2023sam}, SAM-Med2D~\cite{cheng2023sam}, and Medical SAM Adapter~\cite{wu2023medical} introduced Adapter modules and trained the Adapter with medical images. 
They attained good performance on various medical image segmentation tasks. 
MedSAM~\cite{ma2023segment} adapted SAM with more than one million medical image-mask pairs and attained accurate segmentation results. 
MicroSAM~\cite{archit2023segment} also presented a segment anything model for microscopy by fine-tuning SAM with microscopy data. Unlike these approaches that require model fine-tuning for adaptation, our method lifts the blood vessel segmentation capabilities of SAM from 2D images to 3D volumes without any model fine-tuning.

\begin{figure*}[t]
\centering
\includegraphics[width=0.8\textwidth]{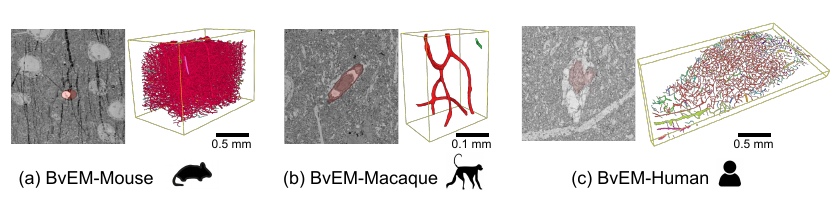}%
\caption{The proposed BvEM dataset. We proofread the blood vessel instance segmentation (displayed in different colors) in the three largest publicly available VEM volumes: (a) mouse~\cite{microns-phase2}, macaque~\cite{loomba2022connectomic}, and human~\cite{shapson2021connectomic} samples acquired at different VEM labs. }
\label{fig:dataset}
\end{figure*}

\section{BvEM Dataset}

\subsection{Dataset Description}
\bfsection{Tissue samples}
We built the BvEM dataset upon the largest publicly available VEM image volumes for three mammal species: visual cortex from an adult mouse~\cite{microns-phase2}, superior temporal gyrus from an adult macaque~\cite{loomba2022connectomic}, and temporal lobe from an adult human~\cite{shapson2021connectomic}. Each volume was acquired with different protocols at different facilities and we refer to respective papers for more details.
As shown in Fig.~\ref{fig:dataset}, the imaging quality and the appearance of blood vessels vary drastically across these three volumes, showcasing the cutting-edge large-scale VEM imaging pipelines.

\bfsection{Image volumes}
We processed the original image volumes into a standardized form that is suitable for benchmarking.
We first downsampled all three VEM image volumes to a near-isotropic resolution ($\sim$200-300 nm) along each dimension, which is a good balance between rich image details for biological analysis and the nature of capillary diameter. 
Then, we trimmed the image boundary of the BvEM-Mouse and BvEM-Human volumes, where the blood vessels are hard to annotate due to the missing image content. 

\subsection{Dataset Annotation} 
\bfsection{Initial annotation}
The paper on the BvEM-human volume provides a manual proofread blood vessel segmentation, where many segments are incomplete or disconnected due to missing annotations.
The papers for BvEM-mouse and BvEM-macaque volumes only provide dense 3D instance segmentation and we manually selected the blood vessel segments.
Due to the large scale of the BvEM-Mouse and BvEM-Human volumes, we only annotated every 4 slices, where the z-dimension resolution is around 1$\mu$m.

\bfsection{Automatic error detection}
For the false split errors, which are the majority source of error, we computed the skeleton of the segmentation and detected all the skeleton endpoints that do not touch the volume boundary as candidates. 
For the false merge errors, we computed the intersection-over-union (IoU) of 2D segments on neighboring slices and detected areas of small IoU value as candidates.

\bfsection{Manual proofreading}
We used the VAST lite software~\cite{berger2018vast} to accelerate the manual proofreading process by using the provided dense segmentation results as templates. Instead of manually delineating segments manually, proofreaders can coarsely draw or fill the segment mask that is snapped to the detailed boundary in the template. 
Assisted with the 3D blood vessel visualizations, two neuroscience experts proofread each volume in multiple rounds until no disagreement. 


\begin{table}[t]
\caption{Dataset information. Despite the difference in the scale and the geometry of the image volume, the largest blood vessel instance (rendered in red) is significantly bigger than the rest combined. The last column shows the max/total length of the blood vessel instances.} 
\centering
\renewcommand\tabcolsep{4pt}
\begin{tabular}{lccc}
\hline
Sample & Resolution (nm) & Size (voxel) & Length: (mm) \\
\hline
Mouse (TEM) & 320$\times$ 256$\times$256 & 2495$\times$3571$\times$2495 & 1.6/1.7\\
Macaque (SBEM) & 240$\times$176$\times$176& 450$\times$1271$\times$995 & 713.3/714.5\\
Human (MultiSEM)&  264$\times$256$\times$256 &  661$\times$7752$\times$13500 &107.2/126.7\\
\hline
\end{tabular}
\label{table:dataset}
\end{table}

\subsection{Dataset Statistics}
\begin{figure}[t]%
\centering
\includegraphics[width=0.8\linewidth]{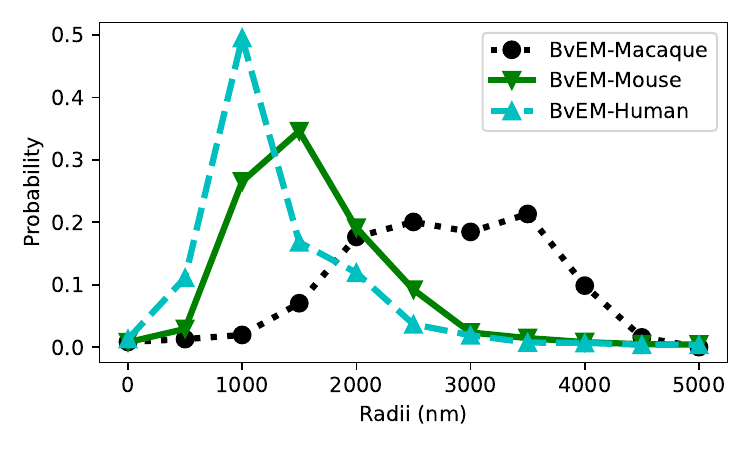}%
\caption{Dataset statistics. Empirical distribution of blood vessel radius.}
\label{fig:dataset_stats}%
\end{figure}

As shown in Tab.~\ref{table:dataset},
the BvEM-Macaque volume has around 0.5G voxels, and the BvEM-Mouse and BvEM-Human volumes are around 80 and 121 times bigger, respectively.
From the blood vessel instance segmentation annotation, we automatically extracted skeleton centerlines~\cite{silversmith2021kimimaro} and computed the length for each blood vessel instance.
Due to the hyper-connectivity nature of cortical blood vessels, the length of the largest instance is around 99\%, 95\%, and 85\% for each volume. 
The histogram of the blood vessel radius in the proposed dataset is shown in Figure~\ref{fig:dataset_stats}.
The mouse and the human dataset have similar radius distributions peaked around capillaries, while the macaque volume has no capillaries.


\begin{figure*}[t]%
\centering
\includegraphics[width=0.9\textwidth]{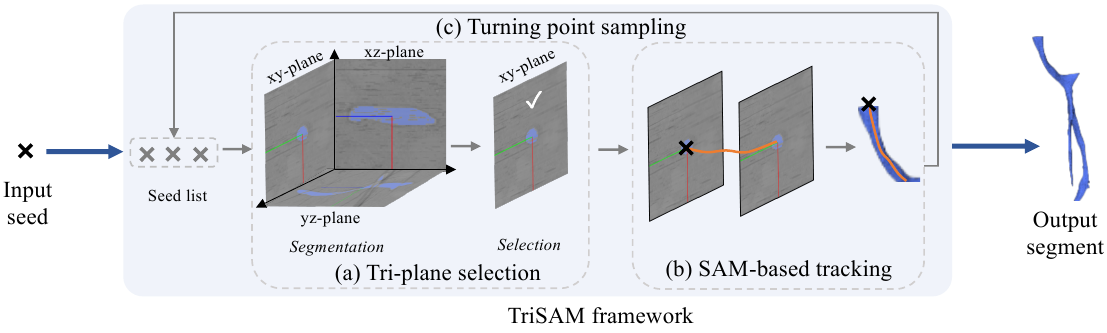}%
\caption{TriSAM framework. (a) Tri-plane selection is first proposed to select the best plane for tracking. (b) SAM-based tracking leverages SAM to perform short-term tracking given a seed location and a tracking axis. (c) Recursive seed sampling exploits potential turning points for long-term tracking. }
\label{fig:pipeline}%
\end{figure*}

\section{Method}\label{sec:method}

\subsection{Overview} 
We aim to leverage the 2D generalist Segment Anything Model (SAM) to build a zero-shot 3D VEM blood vessel segmentation method, which can be widely applied to various image volumes. Given an input seed position, existing SAM-based tracking methods work well when the blood vessel travels only in one axis. However, in reality, blood vessels not only turn in alternating directions but also bifurcate, leading to drastic shape and scale changes.

Instead, we represent the whole connected blood vessel segment that contains the input seed as a graph, $G=(V, E)$, where each node $v\in V$ is a tubular branch segment that does not bifurcate. Thus, the segmentation problem can be solved with the \textbf{graph traversal} algorithm. After \textbf{visiting one node} (\ie, running SAM-based tracking methods given the seed), we sample new seed positions to \textbf{find neighboring nodes} to segment recursively. 

Thus, the proposed TriSAM framework has the following three modules:
(1) \textit{Tri-plane selection module} to pick the best 2D plane to segment the object (Fig.~\ref{fig:pipeline}a);
(2) \textit{SAM-based tracking module} to predict 2D masks along the selected axis (Fig.~\ref{fig:pipeline}b);
(3) \textit{Turning point sampling module} to propose potential turning points as new seeds based on existing segmentation results (Fig.~\ref{fig:pipeline}c) to recursively grow all branches of the blood vessel.
Note that the second module can be any existing SAM-based tracking method.

\subsection{TriSAM Framework} 
We first explain the three modules in the TriSAM framework (Fig.~\ref{fig:pipeline}) and then provide the algorithmic description (Alg.~\ref{alg:trisam}).
In practice, we sample many initial seeds, run the TriSAM framework for each seed, and fuse the segmentation results.

\bfsection{Tri-plane selection module} 
Unlike videos, 3D VEM image volumes $I$ can be segmented on 2D planes along different axes, $p\in\{x,y,z\}$, based on the object morphology. 
However, tracking along some axis $p$ can be harder than others due to the irregular shape of the 2D cross-section of the 3D blood vessels on the corresponding 2D plane $I_p$. 
We later verify such a phenomenon empirically as shown in Table~\ref{table:ablation_plane} in which tracking along the $y$ axis is more effective than tracking along the $x$ axis for a given seed position.

This module aims to select the best axis $p^{*}$ for tracking, given the SAM segmentation results, $\mathrm{SAM}(I_p, s)$, on each tri-plane $I_p$ centered at the seed position $s$ as the approximation of the blood vessel cross-sections.
In theory, the ideal axis for tracking is along the blood flow direction, where the corresponding 2D image plane is tangent to the blood flow direction with the smallest cross-section segment area (Fig.~\ref{fig:pipeline}a, xy-plane).
In addition, SAM outputs the probability, $P_{\mathrm{SAM}}$, for the segment result, reflecting the confidence in the naturalness of its shape.
Thus, to combine the two cues above, we pick the axis with the smallest segment size on the 2D plane and a probability of at least the threshold $\tau$ (\textsc{plane-select} in Alg.~\ref{alg:trisam}).
\begin{align}
p^{*} = \arg\min_p \mathrm{Area}(\mathrm{SAM}(I_p, s)),~  \mathrm{s.t.}~P_{\mathrm{SAM}}(I_p, s) > \tau.
\end{align}

To avoid an infinite recursion, traditional graph traversal algorithms check if a node is marked visited. In our case, a tubular branch segment may grow into different blood vessel segments along different tracking axes, \eg, bifurcation regions.
Thus, we mark the combination of the seed and the tri-plane tracking axis $(s, p^*)$ visited if seed $s$ falls in the segment that is predicted through tracking along $p^*$ axis (\textsc{Visited} in Alg.~\ref{alg:trisam}).
Note that $p^{*}=\phi$ if no plane is selected and \textsc{Visited}($(s, p^*)$) is True.
To implement the ``\textsc{Visited}" function, we store both the segmentation result and the $p^*$, which is omitted in Alg.~\ref{alg:trisam} for simplicity.

\bfsection{SAM-based tracking module}
Given the seed position $s$, the selected tracking axis $p^{*}$ and SAM's initial 2D segment mask, we need to produce a 3D segment by tracking the 2D mask with SAM in both directions (\eg, $x+$ and $x-$).
Note that naively propagating the mask center or bounding box as the SAM prompt for the next image slice leads to poor segmentation results, as SAM may output segments with inconsistent sizes or shrinking sizes along the axis respectively.
Many existing sophisticated SAM-based tracking methods can be directly used.

This module aims to provide a simple yet effective approach by generating better SAM prompts for the next image slice.
Empirically, we find prompting SAM with both the enlarged bounding box and the center of the segment generates better segment tracking results (\textsc{SAM-Track} in Alg.~\ref{alg:trisam}).
\begin{align}
prompt = \{(x,y), (x,y,\gamma w,\gamma h)\},
\end{align}
where $(x,y,\gamma w,\gamma h)$ is the bounding box of the segmentation and $\gamma$ is the scaling factor. 
To ensure the quality of the predicted SAM mask, the proposed tracking module terminates when the SAM probability for the predicted mask is lower than the threshold $\tau$.

\bfsection{Turning point sampling module}
As discussed above, SAM-based tracking module can not handle blood vessel segments that change directions or bifurcate, where the 2D segment along the original direction changes drastically.
Thus, when the SAM-based tracking module terminates in the middle of the volume, we need to find \textbf{turning points} as new seeds to track segments along other directions. 
Naively, we can densely sample points from the segmentation result and run the \textsc{plane-select} module to find points that prefer other directions. However, such an approach can be inefficient as most points may prefer the original tracking direction, and ineffective due to false turning points caused by SAM errors. 

Instead, we design this module to sample turning points around the last point prompt position from the SAM-based tracking module.
We first predict 2D SAM segmentation $seg_{p}$ along the other two directions ($p\in\{x,y,z\}\setminus\{p^*\}$) at the last point prompt position, which has a high probability of capturing the tangent cross-section of the swerving blood vessel. 
Then, we sample $K$ points from each of the 2D segmentation $seg_p$ with the Farthest Point Sampling (FPS) method as turning points added to the seed list.

\begin{align}
seeds = \bigcup_{p\in\{x,y,z\}\setminus\{p^*\}} \mathrm{FPS}(seg_{p}, K).
\end{align}

\begin{algorithm}[t]
\caption{TriSAM Framework}\label{alg:trisam}
\begin{algorithmic}
\Require 3D image volume $I$ and initial seed $s_0$, threshold $\tau$
\State Initialize the prediction: $\tP=\phi$
\State Initialize the seed list: $\tS = \{s_0\}$
\While{ $\tS\neq \phi$}
\State Take $s$ from $\tS$

\State $p^{*}=\textsc{plane-select}(I, s)$ 
\IfNot{$\textsc{Visited}(s, p^{*})$}
\State $seg=\textsc{SAM-Track}(I, s, p^{*})$ 
\If{$seg\neq\phi$}
\State $seeds = \textsc{TurningPoint-Sample}(seg, p^*)$ 
\State $\tP = \tP \cup seg$ 
\State $\tS = \tS \cup seeds$ 
\EndIf
\EndIf
\EndWhile
\State \Return $\tP$
\end{algorithmic}
\end{algorithm}
\bfsection{TriSAM Framework Algorithm}
Given the initial seed $s_0$, the proposed TriSAM framework applies the graph traversal algorithm to segment the whole connected blood vessel segment (Alg.~\ref{alg:trisam}).
As either breadth-first-search or depth-first-search works, we use a generic ``list" data structure ($\tS$) to store the seeds to visit.
At each step, we take out a seed $s\in\tS$, predict the tubular branch with the \textsc{Plane-select} and \textsc{sam-track} modules (\ie, visit a node), and then find its neighboring branches to segment with the \textsc{turningpoint-sample} module (\ie, add neighbors as new seeds).
The TriSAM framework runs recursively until the seed list $\tS=\phi$.




\subsection{Implementation Details} 
\bfsection{Image Pre-processing}
The large-scale datasets have many image artifacts and missing slices, which degrade the SAM-based tracking results.
We utilize the temporal smoothing method along the z-axis to deflicker the images.

\bfsection{Initial seed generation}
Initial seeds can be effectively generated with global color thresholding since the pixels of blood vessels are brighter than the background. 
To improve efficiency, we only keep the center of each connected component as the final seeds.  

\bfsection{SAM details}
Unless stated otherwise, we use MobileSAM \cite{zhang2023faster} instead of the standard SAM \cite{kirillov2023segment} in all conducted experiments to improve the inference speed. 
To clean up SAM results, holes are filled and small connected components are removed through binary morphological operations. 

\bfsection{Hyperparamter selection}
As a zero-shot approach, TriSAM sets the hyperparameters to reasonable values instead of using ground truth data. In later ablation studies, we verify the robustness of hyperparameters within the reasonable range.
For the global color thresholding for seed generation, $\eta$ is set to 98 percentile. 
For SAM-based tracking, the probability threshold $\tau$ is set to 0.8. 

\bfsection{Inference}
For BvEM-Macaque, we directly predict the whole volume. 
For others, $k\times1024\times1024$ subvolumes are used since 1024$\times$1024 is the default resolution for SAM where $k=661$ for BvEM-Human and $k=818$ for BvEM-Mouse. 
The subvolume results are later fused to form the final prediction. 
All experiments are conducted on an NVIDIA-A100 GPU.

\section{Experiments}\label{sec:exp}

\subsection{Experimental Settings}

\bfsection{Evaluation Metrics} We use the Precision, Recall, and Accuracy metric defined in~\cite{weigert2020star}, where $Accuracy = \frac{TP}{TP+FP+FN}$. 
{\small
\begin{align}
    Precision &=\frac{TP}{TP+FP}, \\
    Recall &=\frac{TP}{TP+FN}, \\
    Accuracy &= \frac{TP}{TP+FP+FN},
\end{align}}
where $TP$, $FP$, and $FN$ are instance-level true positive, false positive, and false negative respectively.
We use instance-level metrics since it is more sensitive to split errors. In particular, the Hungarian algorithm is used to match ground-truth instances and predicted instances with negative Accuracy as the cost matrix.
We used the whole dataset for evaluation and computed the score on the largest instance segment of each volume due to its dominating size.

\begin{table*}[t]
\caption{Benchmark results on the proposed BvEM dataset. We evaluate the initial blood vessel annotation to show the significant amount of our proofreading effort. }
\centering
\resizebox{0.9\textwidth}{!}{%
\begin{tabular}{lcccccccccc}
\hline
\multirow{2}{*}{Method} & \multirow{2}{*}{Setting} & \multicolumn{3}{c}{BvEM-Mouse}  & \multicolumn{3}{c}{BvEM-Macaque} & \multicolumn{3}{c}{BvEM-Human} \\
~ & ~ & Pre & Rec & Acc & Pre & Rec & Acc & Pre & Rec & Acc \\
\hline
Initial annotation & N/A &{93.74}&36.62&35.74&1.65&23.17&1.57&{100.00}&25.68&25.68 \\
\hline
Color thresholding & N/A &  \textbf{86.45}& 37.32& 35.26& \textbf{95.14}& 21.65& 21.42  &\textbf{41.77} &1.92 & 1.87  \\
MAESTER \cite{xie2023maester} & unsupervised  &2.16&18.95&0.94 &22.08&40.30&16.64 &0.29&5.03&0.27 \\
3D UNet \cite{microns-phase2} & supervised  &13.45  &0.91  &0.67 & 16.56& 86.95& 16.16&  68.63& 2.46& 2.43\\
nnUNET \cite{isensee2021nnu} & supervised  &3.54&49.55&6.59 &24.34&29.57&23.74 &3.44&23.20&5.86 \\
\hline
SAM+IoU tracking\cite{kirillov2023segment}& zero-shot &  63.59& 0.27& 0.27 &74.39 &1.89 &1.88 &18.19 &23.58  &12.92  \\ 
\textbf{TriSAM} (ours) &zero-shot& 84.12 & \textbf{66.75} & \textbf{59.28} & {78.41} & \textbf{74.97} & \textbf{62.14} & 31.35 & \textbf{25.57} & \textbf{16.39} \\
\hline
\end{tabular}
}
\label{table:main_results}
\end{table*}
\begin{figure*}[t]%
\centering
\includegraphics[width=0.9\textwidth]{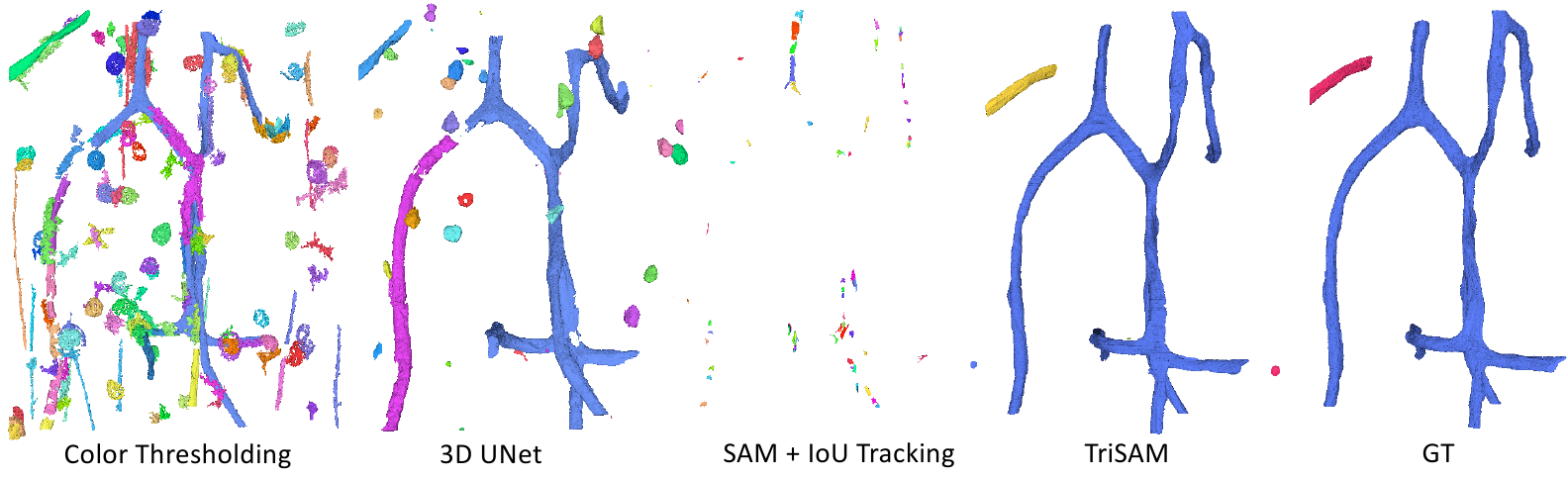}%
\caption{Qualitative instance segmentation results on the BvEM-Macaque volume. Different colors indicate different instances. Color thresholding and 3D UNet often produce false positives, whereas SAM+IoU tracking tends to miss a significant portion of blood vessels. Among the comparison methods, TriSAM segmentation stands out as the most effective.
}
\label{fig:bench}%
\end{figure*}
\subsection{Benchmark Results}

\bfsection{Methods in Comparison}
We compare the proposed TriSAM with both zero-shot baselines and supervised methods. 
The compared zero-shot methods are global color thresholding and SAM+IoU tracking. 
For color thresholding, we first perform (3D) Gaussian blurring with $\sigma=1$ on 3D chunks (10$\times$512$\times$512). Then we label all voxels that are 3 standard derivations above mean as positive.
Finally, connected components with less than 1000 voxels are filtered out.
For SAM+IoU Tracking, we segment all objects in each z-slice of the dataset using automatic mask generation. Then we track each blood vessel using the first labeled slice as seeds. Our simple tracking algorithm finds the mask in the next slice with maximum IoU with the current slice. If the max IoU is above a threshold, we assign this mask to the current object and continue tracking. We also tried SAM+IoU tracking with microSAM \cite{Archit2023SegmentAF} weights that have been finetuned on EM images. This model however does not work well on our dataset. We expect this is because microSAM has been finetuned on high-resolution EM images and does not generalize to our low-resolution dataset.
We further compare TriSAM with the supervised method 3D U-Net \cite{microns-phase2}. We use the implementation from \cite{10.7554/eLife.57613}. 
For supervised methods, we cropped subvolumes at the center of each volume composing approximately 10\% in size. These annotated subvolumes were divided into a 1-1 train-val split.

\bfsection{Results Analysis} 
The results are shown in Tab.~\ref{table:main_results}. First, both the Color Thresholding and SAM+IoU Tracking methods exhibit significant performance variability across three volumes, highlighting the diversity of our dataset and the sensitivity of these methods to different species. Furthermore, both of these unsupervised methods demonstrate relatively poor performance, underscoring the challenges of the zero-shot setting in the BvEM dataset. Additionally, the 3D UNet, as a supervised learning approach, also yields subpar results, indicating poor generalization of models trained with limited data. Finally, TriSAM significantly outperforms other methods as it not only accurately segments the boundary but also tracks the blood vessels in the long term.

\bfsection{Qualitative Results} The final instance segmentation results on BvEM-Macaque are shown in Fig.~\ref{fig:bench}. Color thresholding segments bright pixels, inadvertently capturing nuclei cells while overlooking darker pixels corresponding to blood vessels. Training the 3D UNet model with limited data results in confusion with background elements. IoU tracking fails to capture a significant portion of the blood vessel, revealing its ineffectiveness in tracking. TriSAM prediction emerges as the most accurate method, affirming its effectiveness.

\begin{figure*}[t]%
\begin{minipage}{.32\textwidth}
\centering
\includegraphics[width=0.62\textwidth]{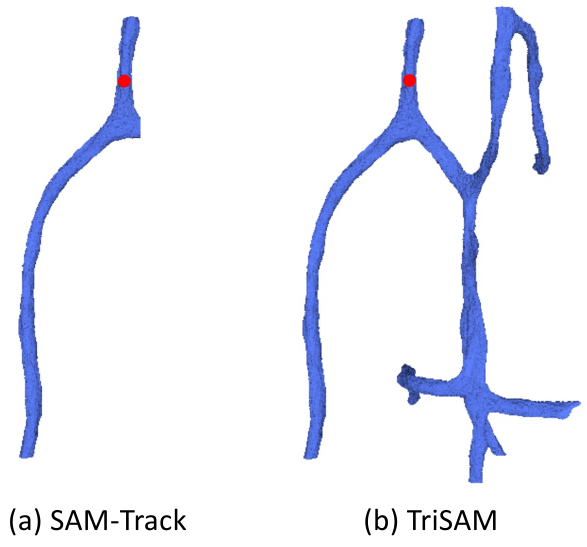}
\caption{Results with oracle 2D segmentation.}
\label{fig:oracle}%
\end{minipage}
\quad
\begin{minipage}{.65\textwidth}
\centering
\includegraphics[width=\textwidth]{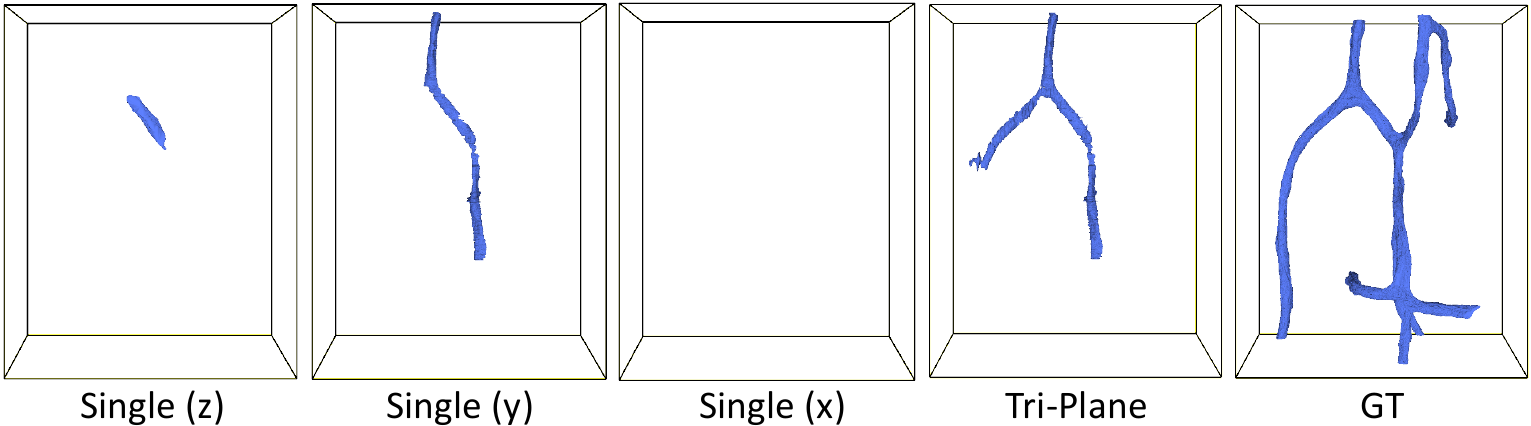}%
\caption{Results for different plane selection strategies without recursive seed sampling.}
\label{fig:visualization}%
\end{minipage}
\end{figure*}

\begin{table}[t]
\caption{Ablation studies on different plane selection strategies. The proposed tri-plane approach achieves the best overall accuracy with comparable speed.} 
\centering
\begin{tabular}{lcccc}
\hline
Method & Pre. (\%) & Rec. (\%)& Acc. (\%) & Speed (sec) \\
\hline
Single-plane (z)   & 75.28 & 48.48 & 41.82 & 324\\
Single-plane (y)   & \textbf{79.37} & 58.47 & 50.75 &\textbf{307} \\
Single-plane (x)   & 69.73 & 12.14 & 11.53 &345 \\ 
Single-plane (fusion) &71.78&74.15& 57.41 & 976\\
\textbf{Tri-plane} &{78.41}&\textbf{74.97}&\textbf{62.14} &335\\
\hline
\end{tabular}
\label{table:ablation_plane}
\end{table}

\subsection{Ablation Studies}

We conducted a comprehensive series of ablation studies exclusively using the BvEM-Macaque dataset due to the computation constraints.

\bfsection{Oracle analysis} 
We conduct an oracle analysis to showcase the advantage of the proposed TriSAM framework for its capability to deal with complicated vascular geometry. 
For the SAM-based tracking module in TriSAM, we plug in an oracle 2D segmentation method which returns the connected component of the ground truth mask containing the prompt. As shown in Fig.~\ref{fig:oracle}b, the TriSAM framework can perfectly segment the whole vasculature with the oracle 2D segmentation. 
In comparison (Fig.~\ref{fig:oracle}a), existing SAM-based video object tracking methods can only track along one axis, missing bifurcated branches.



\bfsection{Tri-plane selection module} We first compare our method with the single-plane methods to evaluate the effectiveness of the tri-plane.
For the single-plane method, we choose one plane as the main plane and only track along the chosen plane. The results on the BvEM-Macaque volume are shown in Tab.~\ref{table:ablation_plane}.
We have observed significant differences in performance among the three single-plane methods, with accuracy ranging from 11.53\% to 50.75\%, depending on the chosen tracking plane, which indicates the importance of the chosen tracking plane. This variability can be attributed to the tubular nature of blood vessel extensions within biological organisms, resulting in the generation of intricate mask shapes in certain planes, while simpler mask shapes are produced in others. Then we fuse the results of three Single-Plane methods and attain a higher accuracy of 57.41\%, which demonstrates that the segmentation results from different planes exhibit a high degree of complementarity with each other. Instead, 
We see that tri-plane selection exploits the blood vessel 3D structures
by tracking along a suitable plane 
and attains the highest accuracy of 62.14\%.

\begin{figure*}[t]
    \centering
    \begin{minipage}{0.3\textwidth}
        \centering
        \includegraphics[width=\textwidth]{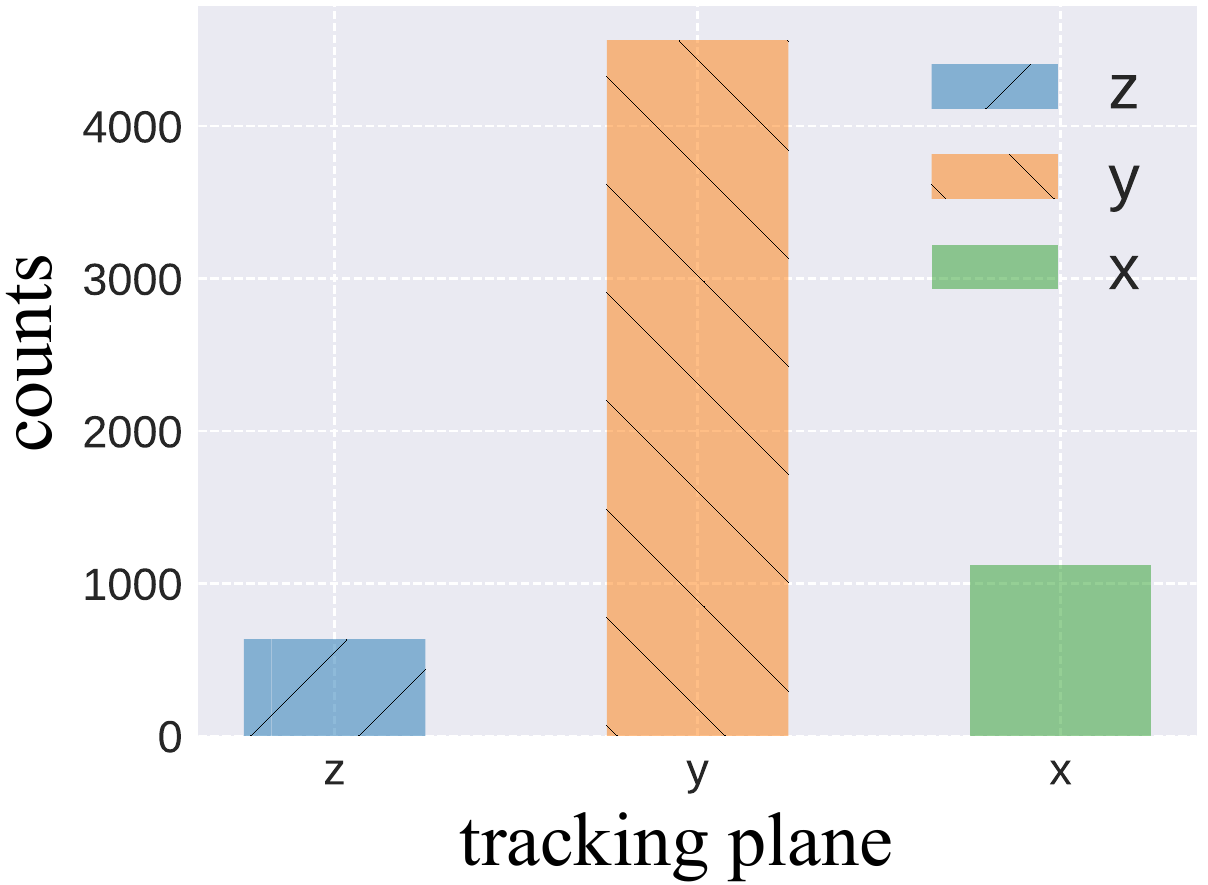} 
        \caption{Histogram of the tri-plane selection.}
        \label{fig:plane_hist}
    \end{minipage}
    \quad
    \begin{minipage}{0.3\textwidth}
        \centering
        \includegraphics[width=\textwidth]{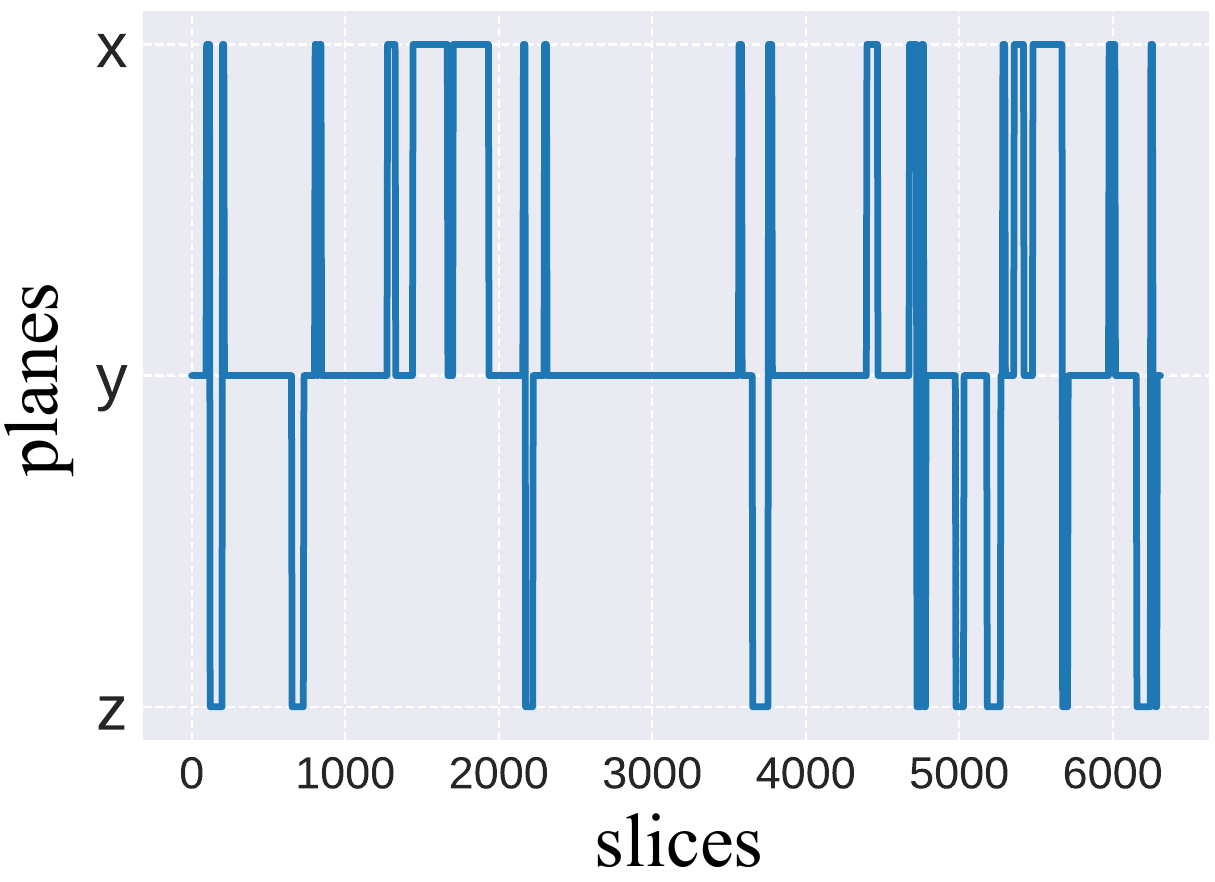} 
        \caption{Dynamics of the tri-plane selection.}
        \label{fig:plane_dynamic}
    \end{minipage}
    \quad
    \begin{minipage}{0.3\textwidth}
        \centering
        \includegraphics[width=\textwidth]{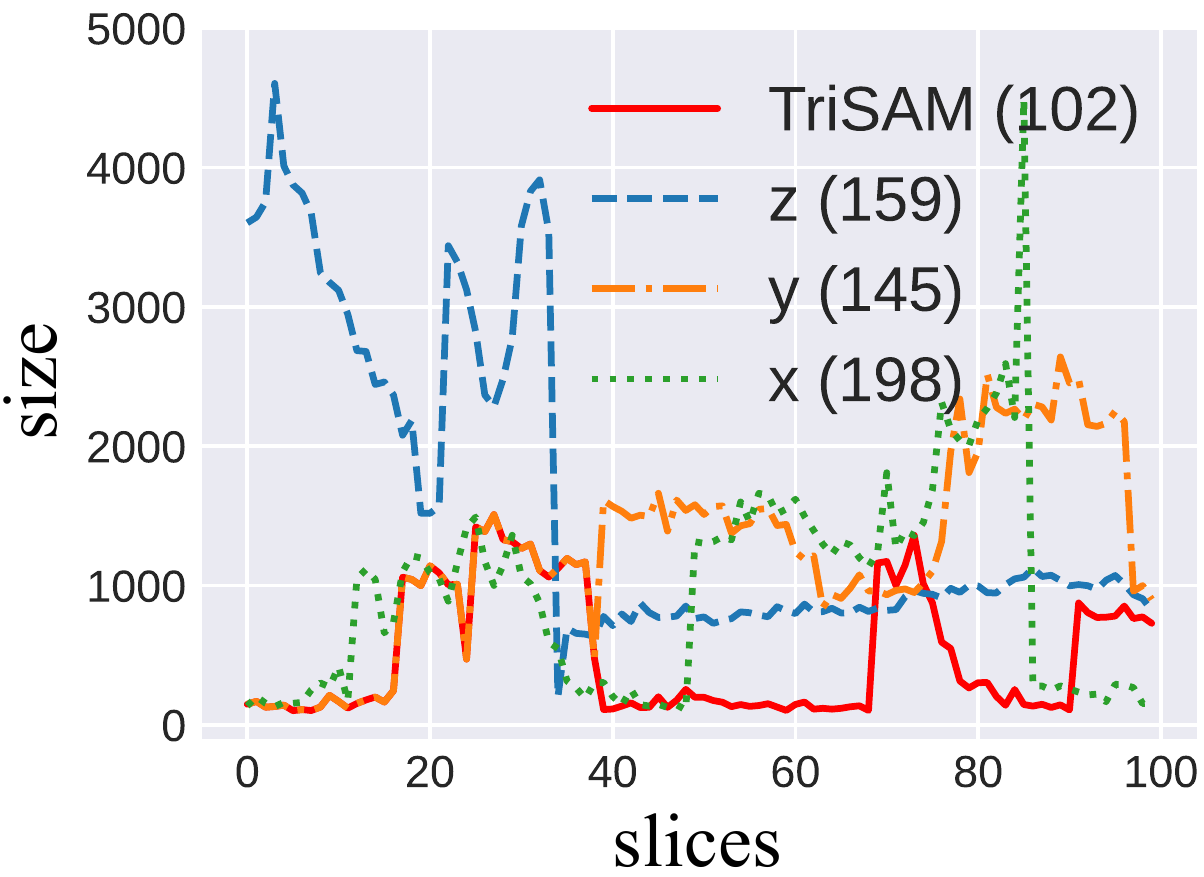} 
        \caption{Dynamics of segment size. 
        }
        \label{fig:size_dynamic}
    \end{minipage}
\end{figure*}

We visualize the segmentation results with one initial seed on BvEM-Macaque in Fig.~\ref{fig:visualization}. 
The origin point is the seed location. 
We see that the performance is sensitive to the selection of the tracking plane. If the plane is not well-selected, the segmentation result can be empty as the example of tracking along the x-axis shows.
The best result is tracking along the y-axis which is still worse than the proposed method since it considers potential turning points to leverage the 3D structure. 
The best result is tracking along the y-axis. However, it still falls short of the proposed method's performance, as the latter takes into account potential turning points to exploit the 3D structure and perform long-term tracking.

\begin{table}[t]
\renewcommand\tabcolsep{5pt}
\caption{Ablation study results on the choice of SAM models. The MobileSAM model achieves better performance with faster inference speed and smaller model size.} 
\centering
\begin{tabular}{lccccc}
\hline
Backbone & Pre (\%) & Rec (\%)& Acc (\%)& Speed & Parameters \\
\hline
SAM \cite{kirillov2023segment}&77.65 &66.03&55.48 & 1535s & 615M \\
MobileSAM \cite{zhang2023faster}  &\textbf{78.41}&\textbf{74.97} & \textbf{62.14}   & \textbf{335s} & \textbf{9.66M} \\
\hline
\end{tabular}
\label{table:ablation_SAM}
\end{table}

\bfsection{SAM-based tracking module}
We examine the effect of different SAM variants, \ie, MobileSAM, on the TriSAM performance.
As shown in Tab.~\ref{table:ablation_SAM}, the inference time of MobileSAM is 22\% of the original SAM which confirms that MobileSAM significantly improves the inference speed. Moreover, the performance of MobileSAM is even better than the original SAM, possibly because the distilled small model is less prone to overfitting to the original natural image domain.

\bfsection{Turning point sampling module} To perform long-term tracking and fully leverage the 3D blood vessel structure, we introduced recursive seed sampling by considering potential turning points. To validate its effectiveness, we report the results on the BvEM-Macaque volume in Tab.~\ref{table:ablation_strategy}
where the runtime comparison for segmentation prediction on the entire BvEM-Macaque data is also included.
Compared to the baseline without recursive seed sampling ``naive'' and dense seed sampling ``dense'', the proposed method achieves the best performance with less than 10\% running time increase.
One naive baseline is to remove the recursive seed sampling component and not consider any potential turning points. This strategy is simple and fast but it fails to exploit the 3D shape prior leading to poor performance. Another strategy is to select the best plane for every tracking step/slice, which densely performs SAM segmentation on each step across three planes. 
Therefore, the performance is limited. 
The proposed method achieves the best performance with less than 10\% running time increase on the BvEM-Macaque dataset.  
Unfortunately, it significantly increases the computation cost. 
Compared to our method, DenseSAM's running time is 5.97 times longer since it needs to segment 3 planes for every tracking step. 
However, we were surprised to observe that the performance of the Dense Redirection strategy was even worse. This could be attributed to the frequent axis changes potentially leading to the omission of certain parts of the blood vessel and causing splitting errors.

\begin{table}[t]
\caption{Ablation studies on different seed sampling strategies.  The proposed recursive seed sampling approach achieves the best overall accuracy with comparable speed.} 
\resizebox{0.5\textwidth}{!}{%
\centering
\begin{tabular}{lcccc}
\hline
Strategy & Pre. (\%)& Rec. (\%)& Acc. (\%)& Speed (sec) \\
\hline
Naive   & 79.37 & 58.47 & 50.75  &\textbf{307} \\
Dense sampling &\textbf{86.94}&24.84&23.95&2001($\uparrow$ 552\%)  \\
\textbf{Recursive sampling} &78.41&\textbf{74.97}&\textbf{62.14}&335 ($\uparrow$ 9\%)\\
\hline
\end{tabular}
}
\label{table:ablation_strategy}
\end{table}

\bfsection{Tri-plane dynamics}
We delve deeper into the plane dynamics in Fig. \ref{fig:plane_hist} and \ref{fig:plane_dynamic}. The majority of the selected plane predominantly tracks along the y-axis. This observation aligns with the experimental results presented in Tab.\ref{table:ablation_plane}, where it is evident that a single plane tracking with the y-axis outperforms the z-axis and x-axis. This is because the blood vessel flows mainly along the y-axis in the test volume. In Fig. \ref{fig:size_dynamic}, we explore the size dynamics using various methods, where the mean derivation is shown in parentheses. The size variation observed with our proposed method is relatively smaller compared to tracking along the y-axis and significantly less than when tracking along the z-axis and x-axis.

\begin{figure}[t]%
\begin{minipage}{.5\textwidth}
\includegraphics[width=0.5\textwidth]{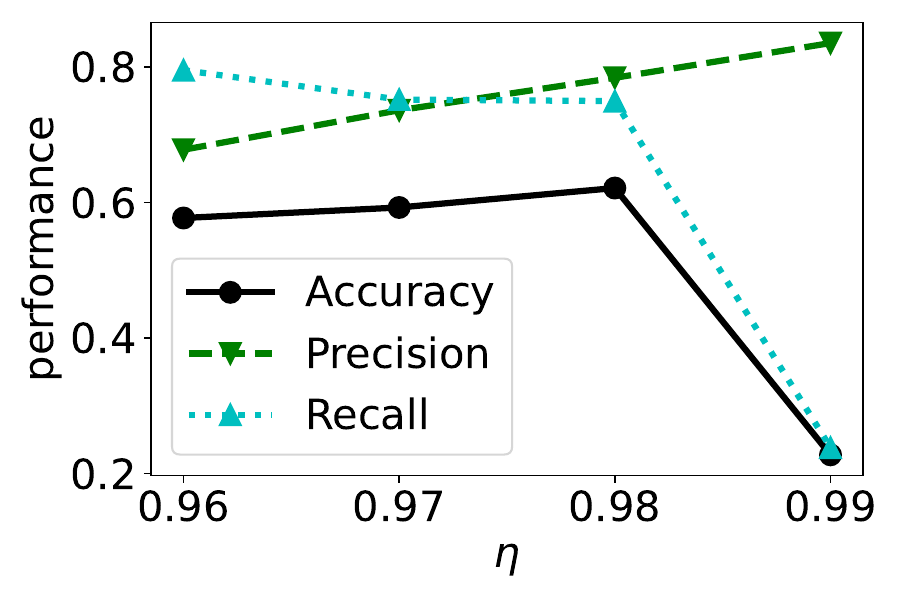}%
\includegraphics[width=0.5\textwidth]{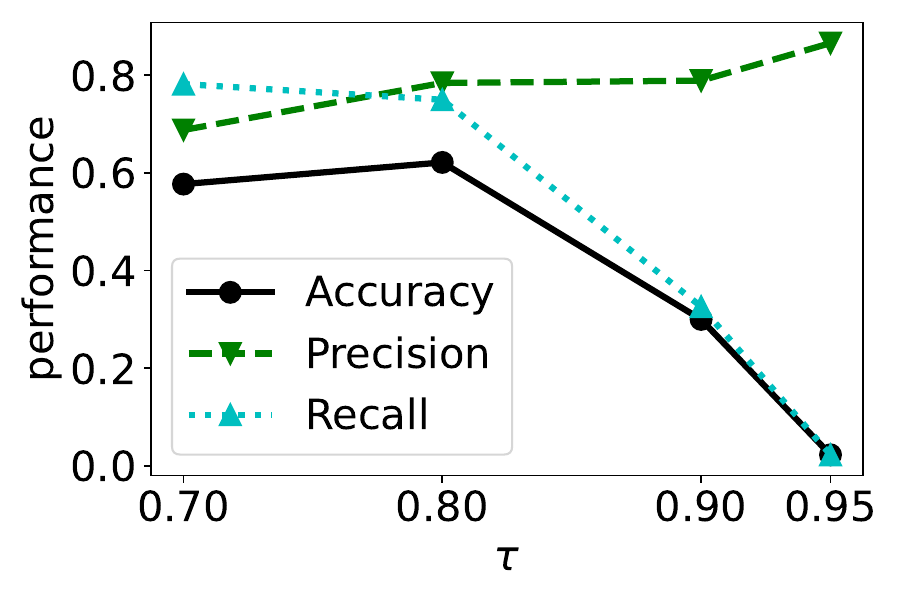}%
\caption{Ablation study results on hyperparameters $\eta$ and $\tau$.}
\label{fig:hyper}%
\end{minipage}
\end{figure}

\bfsection{Hyperparameters} 
We compute the accuracy of TriSAM with varying hyperparameters for seed generation ($\eta$) and SAM-based tracking module ($\tau$).
As shown in Fig. \ref{fig:hyper}, TriSAM's performance is robust within the reasonable range of hyperparameters.

\subsection{Failure Cases}
\begin{figure}[t]%
\centering
\includegraphics[width=0.45\textwidth]{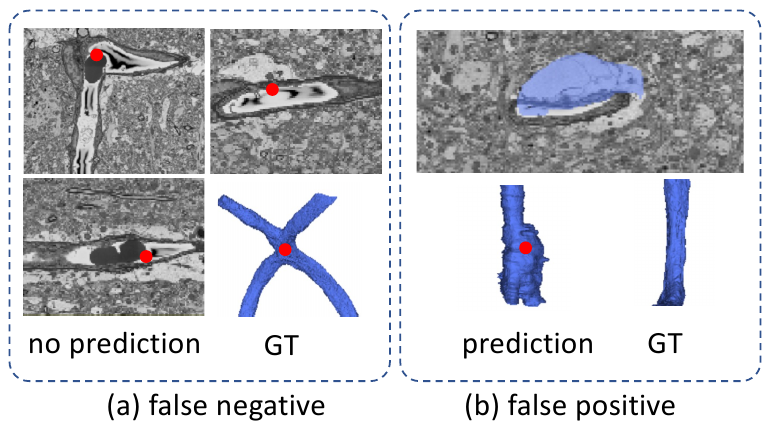}%
\caption{Failure cases of the SAM-based tracking module.}
\label{fig:failures}%
\end{figure}

\textcolor{black}{The main source of error in the proposed TriSAM framework is the SAM-based tracking module, where SAM can not produce accurate 2D mask given the prompt. 
For the false negative case (Fig.~\ref{fig:failures}a), SAM predicts an empty mask when the shape and appearance of the blood vessels are complex (e.g., conjunction point) for all three axes. 
For the false positive case (Fig.~\ref{fig:failures}b), SAM falsely segment the neighboring region with the blood vessel due to the similar image appearance.}

\section{Conclusion}
\label{Conclusion}

\textcolor{black}{In this paper, we contribute the largest-to-date public benchmark, the BvEM dataset, for cortical blood vessel segmentation in 3D VEM images. 
We also developed a zero-shot blood vessel segmentation method, TriSAM, based on the powerful SAM model, offering an efficient and accurate approach for segmenting blood vessels in VEM images. With tri-plane selection, SAM-based tracking, and recursive seed sampling, our TriSAM effectively exploits the 3D blood vessel structure and attains superior performance compared with existing zero-shot and supervised technologies on BvEM across three species, marking a critical step towards unlocking the mysteries of neurovascular coupling and its implications for brain health and pathology. 
With the availability of the BvEM dataset and the TriSAM method, researchers are now equipped with valuable tools to drive breakthroughs in VEM-based cortical blood vessel segmentation and further our understanding of the brain's intricate vascular network.
By addressing a significant gap in the field of neuroimaging, we have laid the foundation for advancing the understanding of cerebral vasculature at the microscale and its intricate relationship with neural function. }

\bibliographystyle{IEEEtran}
\bibliography{trisam.bib}

\end{document}

%% file: math_commands.tex
\usepackage{amsmath,amsfonts,bm}

\def\eqref#1{equation~\ref{#1}}

\def\1{\bm{1}}

\DeclareMathAlphabet{\mathsfit}{\encodingdefault}{\sfdefault}{m}{sl}
\SetMathAlphabet{\mathsfit}{bold}{\encodingdefault}{\sfdefault}{bx}{n}
\newcommand{\tens}[1]{\bm{\mathsfit{#1}}}

\def\tP{{\tens{P}}}

\def\tS{{\tens{S}}}













%% file: tmi.bbl
\begin{thebibliography}{10}
\providecommand{\url}[1]{#1}
\csname url@samestyle\endcsname
\providecommand{\newblock}{\relax}
\providecommand{\bibinfo}[2]{#2}
\providecommand{\BIBentrySTDinterwordspacing}{\spaceskip=0pt\relax}
\providecommand{\BIBentryALTinterwordstretchfactor}{4}
\providecommand{\BIBentryALTinterwordspacing}{\spaceskip=\fontdimen2\font plus
\BIBentryALTinterwordstretchfactor\fontdimen3\font minus
  \fontdimen4\font\relax}
\providecommand{\BIBforeignlanguage}[2]{{%
\expandafter\ifx\csname l@#1\endcsname\relax
\typeout{** WARNING: IEEEtran.bst: No hyphenation pattern has been}%
\typeout{** loaded for the language `#1'. Using the pattern for}%
\typeout{** the default language instead.}%
\else
\language=\csname l@#1\endcsname
\fi
#2}}
\providecommand{\BIBdecl}{\relax}
\BIBdecl

\bibitem{dyer2017quantifying}
E.~L. Dyer, W.~G. Roncal, J.~A. Prasad, H.~L. Fernandes, D.~G{\"u}rsoy,
  V.~De~Andrade, K.~Fezzaa, X.~Xiao, J.~T. Vogelstein, C.~Jacobsen
  \emph{et~al.}, ``Quantifying mesoscale neuroanatomy using x-ray
  microtomography,'' \emph{eneuro}, vol.~4, no.~5, 2017.

\bibitem{lochhead2023high}
J.~J. Lochhead, E.~I. Williams, E.~S. Reddell, E.~Dorn, P.~T. Ronaldson, and
  T.~P. Davis, ``High resolution multiplex confocal imaging of the
  neurovascular unit in health and experimental ischemic stroke,''
  \emph{Cells}, vol.~12, no.~4, p. 645, 2023.

\bibitem{andreone2015neuronal}
B.~J. Andreone, B.~Lacoste, and C.~Gu, ``Neuronal and vascular interactions,''
  \emph{Annual review of neuroscience}, vol.~38, pp. 25--46, 2015.

\bibitem{peters2004selfish}
A.~Peters, U.~Schweiger, L.~Pellerin, C.~Hubold, K.~Oltmanns, M.~Conrad,
  B.~Schultes, J.~Born, and H.~Fehm, ``The selfish brain: competition for
  energy resources,'' \emph{Neuroscience \& Biobehavioral Reviews}, vol.~28,
  no.~2, pp. 143--180, 2004.

\bibitem{iadecola2004neurovascular}
C.~Iadecola, ``Neurovascular regulation in the normal brain and in alzheimer's
  disease,'' \emph{Nature Reviews Neuroscience}, vol.~5, no.~5, pp. 347--360,
  2004.

\bibitem{kalaria2010vascular}
R.~N. Kalaria, ``Vascular basis for brain degeneration: faltering controls and
  risk factors for dementia,'' \emph{Nutrition reviews}, vol.~68, no. suppl\_2,
  pp. S74--S87, 2010.

\bibitem{mcdonald2003imaging}
D.~M. McDonald and P.~L. Choyke, ``Imaging of angiogenesis: from microscope to
  clinic,'' \emph{Nature medicine}, vol.~9, no.~6, pp. 713--725, 2003.

\bibitem{peddie2022volume}
C.~J. Peddie, C.~Genoud, A.~Kreshuk, K.~Meechan, K.~D. Micheva, K.~Narayan,
  C.~Pape, R.~G. Parton, N.~L. Schieber, Y.~Schwab \emph{et~al.}, ``Volume
  electron microscopy,'' \emph{Nature Reviews Methods Primers}, vol.~2, no.~1,
  p.~51, 2022.

\bibitem{goni2022brain}
M.~R. Goni, N.~I.~R. Ruhaiyem, M.~Mustapha, A.~Achuthan, and C.~M. N. C.~M.
  Nassir, ``Brain vessel segmentation using deep learning-a review,''
  \emph{IEEE Access}, 2022.

\bibitem{moccia2018blood}
S.~Moccia, E.~De~Momi, S.~El~Hadji, and L.~S. Mattos, ``Blood vessel
  segmentation algorithms—review of methods, datasets and evaluation
  metrics,'' \emph{Computer methods and programs in biomedicine}, vol. 158, pp.
  71--91, 2018.

\bibitem{microns-phase2}
\BIBentryALTinterwordspacing
{The MICrONS Consortium}, J.~A. Bae, et~al., and C.~Zhang, ``Functional
  connectomics spanning multiple areas of mouse visual cortex,''
  \emph{bioRxiv}, 2023. [Online]. Available:
  \url{https://www.biorxiv.org/content/early/2023/04/19/2021.07.28.454025}
\BIBentrySTDinterwordspacing

\bibitem{shapson2021connectomic}
A.~Shapson-Coe, M.~Januszewski, D.~R. Berger, A.~Pope, Y.~Wu, T.~Blakely, R.~L.
  Schalek, P.~H. Li, S.~Wang, J.~Maitin-Shepard \emph{et~al.}, ``A connectomic
  study of a petascale fragment of human cerebral cortex,'' \emph{BioRxiv}, pp.
  2021--05, 2021.

\bibitem{flywire}
\BIBentryALTinterwordspacing
S.~Dorkenwald, et~al., and the FlyWire~Consortium, ``Neuronal wiring diagram of
  an adult brain,'' \emph{bioRxiv}, 2023. [Online]. Available:
  \url{https://www.biorxiv.org/content/early/2023/07/11/2023.06.27.546656}
\BIBentrySTDinterwordspacing

\bibitem{abbott2020mind}
L.~F. Abbott, D.~D. Bock, E.~M. Callaway, W.~Denk, C.~Dulac, A.~L. Fairhall,
  I.~Fiete, K.~M. Harris, M.~Helmstaedter, V.~Jain \emph{et~al.}, ``The mind of
  a mouse,'' \emph{Cell}, vol. 182, no.~6, pp. 1372--1376, 2020.

\bibitem{kirillov2023segment}
A.~Kirillov, E.~Mintun, N.~Ravi, H.~Mao, C.~Rolland, L.~Gustafson, T.~Xiao,
  S.~Whitehead, A.~C. Berg, W.-Y. Lo \emph{et~al.}, ``Segment anything,''
  \emph{arXiv preprint arXiv:2304.02643}, 2023.

\bibitem{yang2023track}
J.~Yang, M.~Gao, Z.~Li, S.~Gao, F.~Wang, and F.~Zheng, ``Track anything:
  Segment anything meets videos,'' \emph{arXiv preprint arXiv:2304.11968},
  2023.

\bibitem{januszewski2018high}
M.~Januszewski, J.~Kornfeld, P.~H. Li, A.~Pope, T.~Blakely, L.~Lindsey,
  J.~Maitin-Shepard, M.~Tyka, W.~Denk, and V.~Jain, ``High-precision automated
  reconstruction of neurons with flood-filling networks,'' \emph{Nature
  methods}, vol.~15, no.~8, pp. 605--610, 2018.

\bibitem{lee2019convolutional}
K.~Lee, N.~Turner, T.~Macrina, J.~Wu, R.~Lu, and H.~S. Seung, ``Convolutional
  nets for reconstructing neural circuits from brain images acquired by serial
  section electron microscopy,'' \emph{Current opinion in neurobiology},
  vol.~55, pp. 188--198, 2019.

\bibitem{wu2023out}
J.~Wu, Y.~Li, N.~Gupta, K.~Shinomiya, P.~Gunn, A.~Polilov, H.~Pfister,
  D.~Chklovskii, and D.~Wei, ``An out-of-domain synapse detection challenge for
  microwasp brain connectomes,'' \emph{arXiv preprint arXiv:2302.00545}, 2023.

\bibitem{turner2020synaptic}
N.~L. Turner, K.~Lee, R.~Lu, J.~Wu, D.~Ih, and H.~S. Seung, ``Synaptic partner
  assignment using attentional voxel association networks,'' in \emph{2020 IEEE
  17th International Symposium on Biomedical Imaging (ISBI)}.\hskip 1em plus
  0.5em minus 0.4em\relax IEEE, 2020, pp. 1--5.

\bibitem{buhmann2021automatic}
J.~Buhmann, A.~Sheridan, C.~Malin-Mayor, P.~Schlegel, S.~Gerhard, T.~Kazimiers,
  R.~Krause, T.~M. Nguyen, L.~Heinrich, W.-C.~A. Lee \emph{et~al.}, ``Automatic
  detection of synaptic partners in a whole-brain drosophila electron
  microscopy data set,'' \emph{Nature methods}, vol.~18, no.~7, pp. 771--774,
  2021.

\bibitem{tsai2009correlations}
P.~S. Tsai, J.~P. Kaufhold, P.~Blinder, B.~Friedman, P.~J. Drew, H.~J. Karten,
  P.~D. Lyden, and D.~Kleinfeld, ``Correlations of neuronal and microvascular
  densities in murine cortex revealed by direct counting and colocalization of
  nuclei and vessels,'' \emph{Journal of Neuroscience}, vol.~29, no.~46, pp.
  14\,553--14\,570, 2009.

\bibitem{ji2021brain}
X.~Ji, T.~Ferreira, B.~Friedman, R.~Liu, H.~Liechty, E.~Bas, J.~Chandrashekar,
  and D.~Kleinfeld, ``Brain microvasculature has a common topology with local
  differences in geometry that match metabolic load,'' \emph{Neuron}, vol. 109,
  no.~7, pp. 1168--1187, 2021.

\bibitem{tetteh2020deepvesselnet}
G.~Tetteh, V.~Efremov, N.~D. Forkert, M.~Schneider, J.~Kirschke, B.~Weber,
  C.~Zimmer, M.~Piraud, and B.~H. Menze, ``Deepvesselnet: Vessel segmentation,
  centerline prediction, and bifurcation detection in 3-d angiographic
  volumes,'' \emph{Frontiers in Neuroscience}, vol.~14, p. 1285, 2020.

\bibitem{livne2019u}
M.~Livne, J.~Rieger, O.~U. Aydin, A.~A. Taha, E.~M. Akay, T.~Kossen,
  J.~Sobesky, J.~D. Kelleher, K.~Hildebrand, D.~Frey \emph{et~al.}, ``A u-net
  deep learning framework for high performance vessel segmentation in patients
  with cerebrovascular disease,'' \emph{Frontiers in neuroscience}, vol.~13,
  p.~97, 2019.

\bibitem{dang2022vessel}
V.~N. Dang, F.~Galati, R.~Cortese, G.~Di~Giacomo, V.~Marconetto, P.~Mathur,
  K.~Lekadir, M.~Lorenzi, F.~Prados, and M.~A. Zuluaga, ``Vessel-captcha: an
  efficient learning framework for vessel annotation and segmentation,''
  \emph{Medical Image Analysis}, vol.~75, p. 102263, 2022.

\bibitem{cheng2023segment}
Y.~Cheng, L.~Li, Y.~Xu, X.~Li, Z.~Yang, W.~Wang, and Y.~Yang, ``Segment and
  track anything,'' \emph{arXiv preprint arXiv:2305.06558}, 2023.

\bibitem{yu2023inpaint}
T.~Yu, R.~Feng, R.~Feng, J.~Liu, X.~Jin, W.~Zeng, and Z.~Chen, ``Inpaint
  anything: Segment anything meets image inpainting,'' \emph{arXiv preprint
  arXiv:2304.06790}, 2023.

\bibitem{ma2023segment}
J.~Ma and B.~Wang, ``Segment anything in medical images,'' \emph{arXiv preprint
  arXiv:2304.12306}, 2023.

\bibitem{deng2023segment}
R.~Deng, C.~Cui, Q.~Liu, T.~Yao, L.~W. Remedios, S.~Bao, B.~A. Landman, L.~E.
  Wheless, L.~A. Coburn, K.~T. Wilson \emph{et~al.}, ``Segment anything model
  (sam) for digital pathology: Assess zero-shot segmentation on whole slide
  imaging,'' \emph{arXiv preprint arXiv:2304.04155}, 2023.

\bibitem{yao2023matte}
J.~Yao, X.~Wang, L.~Ye, and W.~Liu, ``Matte anything: Interactive natural image
  matting with segment anything models,'' \emph{arXiv preprint
  arXiv:2306.04121}, 2023.

\bibitem{lu2023can}
Z.~Lu, Z.~Xiao, J.~Bai, Z.~Xiong, and X.~Wang, ``Can sam boost video
  super-resolution?'' \emph{arXiv preprint arXiv:2305.06524}, 2023.

\bibitem{liu2023segment}
Y.~Liu, L.~Kong, J.~Cen, R.~Chen, W.~Zhang, L.~Pan, K.~Chen, and Z.~Liu,
  ``Segment any point cloud sequences by distilling vision foundation models,''
  \emph{arXiv preprint arXiv:2306.09347}, 2023.

\bibitem{gao2023editanything}
S.~Gao, Z.~Lin, X.~Xie, P.~Zhou, M.-M. Cheng, and S.~Yan, ``Editanything:
  Empowering unparalleled flexibility in image editing and generation,'' in
  \emph{Proceedings of the 31st ACM International Conference on Multimedia,
  Demo track}, 2023.

\bibitem{zhao2023fast}
X.~Zhao, W.~Ding, Y.~An, Y.~Du, T.~Yu, M.~Li, M.~Tang, and J.~Wang, ``Fast
  segment anything,'' \emph{arXiv preprint arXiv:2306.12156}, 2023.

\bibitem{zhang2023faster}
C.~Zhang, D.~Han, Y.~Qiao, J.~U. Kim, S.-H. Bae, S.~Lee, and C.~S. Hong,
  ``Faster segment anything: Towards lightweight sam for mobile applications,''
  \emph{arXiv preprint arXiv:2306.14289}, 2023.

\bibitem{chen2023sam}
T.~Chen, L.~Zhu, C.~Ding, R.~Cao, S.~Zhang, Y.~Wang, Z.~Li, L.~Sun, P.~Mao, and
  Y.~Zang, ``Sam fails to segment anything?--sam-adapter: Adapting sam in
  underperformed scenes: Camouflage, shadow, and more,'' \emph{arXiv preprint
  arXiv:2304.09148}, 2023.

\bibitem{archit2023segment}
A.~Archit, S.~Nair, N.~Khalid, P.~Hilt, V.~Rajashekar, M.~Freitag, S.~Gupta,
  A.~Dengel, S.~Ahmed, and C.~Pape, ``Segment anything for microscopy,''
  \emph{bioRxiv}, pp. 2023--08, 2023.

\bibitem{gong20233dsam}
S.~Gong, Y.~Zhong, W.~Ma, J.~Li, Z.~Wang, J.~Zhang, P.-A. Heng, and Q.~Dou,
  ``3dsam-adapter: Holistic adaptation of sam from 2d to 3d for promptable
  medical image segmentation,'' \emph{arXiv preprint arXiv:2306.13465}, 2023.

\bibitem{mazurowski2023segment}
M.~A. Mazurowski, H.~Dong, H.~Gu, J.~Yang, N.~Konz, and Y.~Zhang, ``Segment
  anything model for medical image analysis: an experimental study,''
  \emph{Medical Image Analysis}, vol.~89, p. 102918, 2023.

\bibitem{cheng2023sam}
J.~Cheng, J.~Ye, Z.~Deng, J.~Chen, T.~Li, H.~Wang, Y.~Su, Z.~Huang, J.~Chen,
  L.~Jiang \emph{et~al.}, ``Sam-med2d,'' \emph{arXiv preprint
  arXiv:2308.16184}, 2023.

\bibitem{wu2023medical}
J.~Wu, R.~Fu, H.~Fang, Y.~Liu, Z.~Wang, Y.~Xu, Y.~Jin, and T.~Arbel, ``Medical
  sam adapter: Adapting segment anything model for medical image
  segmentation,'' \emph{arXiv preprint arXiv:2304.12620}, 2023.

\bibitem{loomba2022connectomic}
S.~Loomba, J.~Straehle, V.~Gangadharan, N.~Heike, A.~Khalifa, A.~Motta, N.~Ju,
  M.~Sievers, J.~Gempt, H.~S. Meyer \emph{et~al.}, ``Connectomic comparison of
  mouse and human cortex,'' \emph{Science}, vol. 377, no. 6602, p. eabo0924,
  2022.

\bibitem{berger2018vast}
D.~R. Berger, H.~S. Seung, and J.~W. Lichtman, ``Vast (volume annotation and
  segmentation tool): efficient manual and semi-automatic labeling of large 3d
  image stacks,'' \emph{Frontiers in neural circuits}, vol.~12, p.~88, 2018.

\bibitem{silversmith2021kimimaro}
W.~Silversmith, J.~Bae, P.~Li, and A.~Wilson, ``Kimimaro: Skeletonize densely
  labeled 3d image segmentations,'' \emph{Zenodo https://doi.
  org/10.5281/zenodo}, vol. 5539912, 2021.

\bibitem{weigert2020star}
M.~Weigert, U.~Schmidt, R.~Haase, K.~Sugawara, and G.~Myers, ``Star-convex
  polyhedra for 3d object detection and segmentation in microscopy,'' in
  \emph{Proceedings of the IEEE/CVF winter conference on applications of
  computer vision}, 2020, pp. 3666--3673.

\bibitem{xie2023maester}
R.~Xie, K.~Pang, G.~D. Bader, and B.~Wang, ``Maester: Masked autoencoder guided
  segmentation at pixel resolution for accurate, self-supervised subcellular
  structure recognition,'' in \emph{Proceedings of the IEEE/CVF Conference on
  Computer Vision and Pattern Recognition}, 2023, pp. 3292--3301.

\bibitem{isensee2021nnu}
F.~Isensee, P.~F. Jaeger, S.~A. Kohl, J.~Petersen, and K.~H. Maier-Hein,
  ``nnu-net: a self-configuring method for deep learning-based biomedical image
  segmentation,'' \emph{Nature methods}, vol.~18, no.~2, pp. 203--211, 2021.

\bibitem{Archit2023SegmentAF}
\BIBentryALTinterwordspacing
A.~Archit, S.~Nair, N.~Khalid, P.~Hilt, V.~Rajashekar, M.~Freitag, S.~Gupta,
  A.~R. Dengel, S.~Ahmed, and C.~Pape, ``Segment anything for microscopy,''
  \emph{bioRxiv}, 2023. [Online]. Available:
  \url{https://api.semanticscholar.org/CorpusID:261125076}
\BIBentrySTDinterwordspacing

\bibitem{10.7554/eLife.57613}
\BIBentryALTinterwordspacing
A.~Wolny, L.~Cerrone, A.~Vijayan, R.~Tofanelli, A.~V. Barro, M.~Louveaux,
  C.~Wenzl, S.~Strauss, D.~Wilson-Sánchez, R.~Lymbouridou, S.~S. Steigleder,
  C.~Pape, A.~Bailoni, S.~Duran-Nebreda, G.~W. Bassel, J.~U. Lohmann,
  M.~Tsiantis, F.~A. Hamprecht, K.~Schneitz, A.~Maizel, and A.~Kreshuk,
  ``Accurate and versatile 3d segmentation of plant tissues at cellular
  resolution,'' \emph{eLife}, vol.~9, p. e57613, jul 2020. [Online]. Available:
  \url{https://doi.org/10.7554/eLife.57613}
\BIBentrySTDinterwordspacing

\end{thebibliography}
